\documentclass[letterpaper, 10 pt, conference]{ieeeconf}  % Comment this line out if you need a4paper

\IEEEoverridecommandlockouts                              % This command is only needed if 
\usepackage{graphicx}
\usepackage{amsmath} % assumes amsmath package installed
\usepackage{amssymb}  % assumes amsmath package installed
\usepackage{tikz}
\usepackage{standalone}
\usepackage{hyperref}
\usepackage{cite}

\usetikzlibrary{arrows.meta,positioning,shapes.geometric,calc}
\usepackage[caption=false,font=footnotesize]{subfig}
\title{\LARGE \bf
Benchmarking Action Spaces in Reinforcement Learning\\for Vision-based Robotic Manipulation
}

\newcommand{\bs}{\boldsymbol}

\newcommand{\x}{\bs{x}}
\newcommand{\ve}{\bs{v}}
\newcommand{\q}{\bs{q}}

\usepackage{etoolbox}
\makeatletter

\author{
Seyed Alireza Azimi$^{1,*}$\quad
Homayoon Farrahi$^{1}$\quad
Abhishek Naik$^{2}$\quad
Colin Bellinger$^{3,4,\dagger}$\quad
A. Rupam Mahmood$^{1,5,6}$\\[0.5em]
{\small $^{1}$Department of Computing Science, University of Alberta}\\
{\small $^{2}$National Research Council Canada}\\
{\small $^{3}$School of Electrical Engineering and Computer Science, University of Ottawa}\\
{\small $^{4}$Vector Institute}\\
{\small $^{5}$Alberta Machine Intelligence Institute (Amii)}\\
{\small $^{6}$Canada CIFAR AI Chair}\\
{\small $^{\dagger}$This work was conducted while at the National Research Council Canada.}\\
{\small $^{*}$Corresponding author: \texttt{sazimi@ualberta.ca}}\\
{\small Demo video: \url{https://youtu.be/MmXEexVRa18}}\\
{\small Code: \url{https://github.com/RL-Sim-to-Real/training}}
}

\begin{document}

\maketitle
\thispagestyle{empty}
\pagestyle{empty}

%%%%%%%%%%%%%%%%%%%%%%%%%%%%%%%%%%%%%%%%%%%%%%%%%%%%%%%%%%%%%%%%%%%%%%%%%%%%%%%%
\begin{abstract} 
In real-world reinforcement learning (RL), the choice of action space can play a key role in shaping motion smoothness, safety, and overall task performance. 
In this study, we evaluate pose increment, pose velocity, joint position increment, and joint velocity across two vision-based manipulation tasks: object picking and pushing. 
We train policies in simulation and deploy them to the real world using sim-to-real transfer. 
We find that action-space representation indeed significantly affects sim-to-real performance. In particular, we find that the joint velocity action space is best for the vision-based picking and pushing tasks in terms of smoothness and final task performance.
We also provide practical guidance for RL practitioners in choosing action spaces for both simulation and real-world experiments. 
\end{abstract}

%%%%%%%%%%%%%%%%%%%%%%%%%%%%%%%%%%%%%%%%%%%%%%%%%%%%%%%%%%%%%%%%%%%%%%%%%%%%%%%%
\section{Introduction}
Reinforcement learning (RL) is an effective framework for learning control policies for robotic tasks. Despite this, RL has faced challenges with real-world practicality, safety, and motion smoothness in robotic manipulation \cite{MahmoodKorenkevychKomerBergstra2018, mahmood2018benchmarking, PengAndrychowiczZarembaAbbeel2017}. A central design choice in any RL formulation is the action space, which has been shown to affect factors such as motion smoothness, sample efficiency, and sim-to-real transfer performance \cite{aljalbout2024role, Kim2023TorqueAgnostic, martin-martin2019variable}. 

Despite the importance of action-space design, prior work has largely conducted this study without vision-based observations and has often limited its study to simulation \cite{mahmood2018benchmarking, aljalbout2024role, varin2019comparison, martin-martin2019variable}. Vision-based manipulation introduces partial observability and perceptual noise, which can substantially alter learning dynamics and real-world behavior. These challenges are further amplified in sim-to-real transfer, where mismatches in dynamics and sensing can expose sensitivities that remain hidden in idealized simulation environments.

In this work, we investigate the impact of action-space design in vision-based robotic manipulation. Vision provides rich sensory input, enabling robots to perceive and interact with their environment.

We study this problem using two vision-based manipulation tasks on the Franka Emika Panda robot arm: PandaPickCuboid and PandaPushCuboid. PandaPickCuboid is a prehensile manipulation task in which the agent must pick up a red cuboid using visual input from a wrist-mounted camera. Object picking is a fundamental robotic manipulation task \cite{zakka2025mujocoplayground, Xu2024MuJoCoFranka, Levine2018Grasping, Lee2019} that naturally decomposes into reaching, grasping, and lifting, making it both challenging and important. PandaPushCuboid is a non-prehensile manipulation task in which the agent must push the same cuboid within a specified boundary. Pushing is another core manipulation task that enables object displacement without grasping and involves interactions with friction and mass \cite{PengAndrychowiczZarembaAbbeel2017, aljalbout2024role}.

Policies are trained in simulation using PPO and evaluated on a physical Franka Emika Panda robot via sim-to-real transfer. Simulation training allows us to safely exploit the exploratory nature of RL, particularly in contact-rich tasks, while enabling rapid experimentation and large-scale data collection. To mitigate discrepancies between simulated and real-world dynamics and visual appearance, we employ domain randomization \cite{zakka2025mujocoplayground, TobinFongRaySchneiderZarembaAbbeel2017}. We adopt PPO due to its compatibility with vision-based actor–critic architectures \cite{SchulmanEtAl2017PPO, brax2021github}, and demonstrated effectiveness in sim-to-real robotic learning \cite{martin-martin2019variable, aljalbout2024role, Tan2018SimToReal, Chebotar2019ClosingSimToReal, OpenAI2019RubiksCube}. Additionally, PPO’s ability to leverage large numbers of parallel simulation environments makes it time-efficient for training. 

We compare four commonly used action-space representations: pose increment, pose velocity, joint velocity, and joint position increment. Our results show that the choice of action space affects final sim-to-real performance, and that joint velocity performs best in terms of smoothness and final task performance.

\section{Related Work}

Action-space design plays a critical role in reinforcement learning for robotic manipulation, influencing learning efficiency, motion quality, and robustness to sim-to-real transfer. A growing body of work has investigated how different action representations affect policy learning and performance across a variety of robotic tasks and embodiments.

Aljalbout et al. \cite{aljalbout2024role} conducted a comprehensive study of action-space design for sim-to-real robotic manipulation, comparing action representations across joint-space, Cartesian-space, and incremental formulations. Using Proximal Policy Optimization (PPO), they trained policies in simulation on a Franka Emika Panda robot and evaluated them on reaching and pushing tasks, considering metrics such as episodic return, constraint violations, task accuracy, and trajectory error. The authors found that action-space choice significantly affects both learning dynamics and transfer performance, with joint-velocity control achieving the strongest overall sim-to-real results while requiring minimal actuator tuning. Our work builds on these findings by extending the evaluation of action spaces to vision-based manipulation and contact-rich picking and pushing tasks.

Martín-Martín et al. \cite{martin-martin2019variable} examined the effect of action-space design on contact-rich manipulation in simulation, comparing joint torque, joint velocity, joint position, and Cartesian impedance control across path following, door opening, and surface wiping tasks. Using PPO, they demonstrated that action-space choice impacts sample efficiency, energy consumption, and safety, with Cartesian impedance actions performing favorably across tasks. While their study focuses on simulation-only settings with state-based observations, we extend this line of work to real-world evaluation via sim-to-real transfer and vision-based sensing.

Similarly, Varin et al. \cite{varin2019comparison} compared several action representations—including torque control, joint-space proportional–derivative (PD) control, inverse dynamics control, and task-space impedance control—across simulated manipulation tasks such as peg insertion, hammering, and pushing. Policies were trained using both PPO and Soft Actor-Critic (SAC), with task-space impedance control consistently achieving high performance and strong sample efficiency. In contrast to their simulation-focused evaluation, our study evaluates action spaces in the real world under partial observability induced by visual input.

Beyond manipulation, action-space design has also been studied in the context of locomotion and control of articulated figures. Peng et al. \cite{Peng2017ActionSpace} investigated action representations for planar articulated figures in simulation, comparing torque control, musculotendon unit activations, PD target angles, and joint velocity commands. They showed that action spaces incorporating intrinsic feedback—such as PD targets and velocity control—improve learning speed and robustness compared to direct torque control. However, their experiments were limited to two-dimensional simulated systems and did not include real-world evaluation.

Chen et al. \cite{chen2023learning} demonstrated that torque control outperforms joint position control in quadruped locomotion tasks under varying gain parameters, although position-based control exhibited superior sample efficiency. Note that sample efficiency would primarily be a concern if it affects training time. Similarly,     Kim et al. \cite{Kim2023TorqueAgnostic} showed that torque-based policies achieve stronger sim-to-real transfer for bipedal locomotion compared to position-based control, albeit at the cost of reduced sample efficiency. They mitigate this issue through pretraining with gravity compensation and highlight the drawback of position control in requiring careful gain tuning across tasks and embodiments—a challenge we also encountered.

Schneider et al. \cite{Schneider2023ActionRepresentations} studied the effect of action representations on policy gradient methods across several simulated environments. For the Gymnasium Reacher task involving a simple two-joint robotic arm, they found that joint velocity control outperformed both torque and joint position control, with torque control performing the worst. Their results further demonstrate that action-space choice can substantially affect learning outcomes across a range of control tasks, including Pendulum and Walker-Walk \cite{tunyasuvunakool2020}.

\section{Sim-to-Real System overview}
In this section, we describe critical components in our sim-to-real pipeline. In particular, the learning algorithm, simulator, and the real-robot control system.
\subsection{Proximal Policy Optimization}
Proximal Policy Optimization (PPO) is an on-policy deep reinforcement learning method designed on the idea of a clipped surrogate objective that confines policy updates to the boundaries of a trust region \cite{SchulmanEtAl2017PPO}:
\begin{align}
    L^{\text{CLIP}}(\boldsymbol{\theta}) 
    := \mathbb{E}_{\pi}\!\left[\, 
    \min\!\left( 
    \rho_t(\boldsymbol{\theta})\, H_t,\; 
    \text{clip}\!\left(\rho_t(\boldsymbol{\theta}), 1 - \epsilon, 1 + \epsilon\right) H_t 
    \right) 
    \right],
\end{align}
where $\rho_t(\boldsymbol{\theta}) = \frac{\pi(A_t \mid S_t, \boldsymbol{\theta})}{\pi(A_t \mid S_t, \boldsymbol{\theta}_{\text{old}})}$ and $H_t$ is the generalized advantage estimate.

We utilize PPO for three reasons: (1) it is a deep reinforcement learning algorithm well suited to continuous state and action spaces, as well as high-dimensional observations such as images; (2) it is widely adopted in prior sim-to-real work \cite{martin-martin2019variable, aljalbout2024role, Tan2018SimToReal, Chebotar2019ClosingSimToReal, OpenAI2019RubiksCube}; and (3) it is time efficient as it can leverage a large number of parallel environments and perform simple updates \cite{SchulmanEtAl2017PPO}.
In our experiments, we use PPO implemented by \cite{brax2021github}.

\subsection{Dynamics-based Simulation}
\begin{figure}[t]
  \centering
  \includegraphics[
    width=\linewidth,
    trim=10 100 10 150,
    clip
  ]{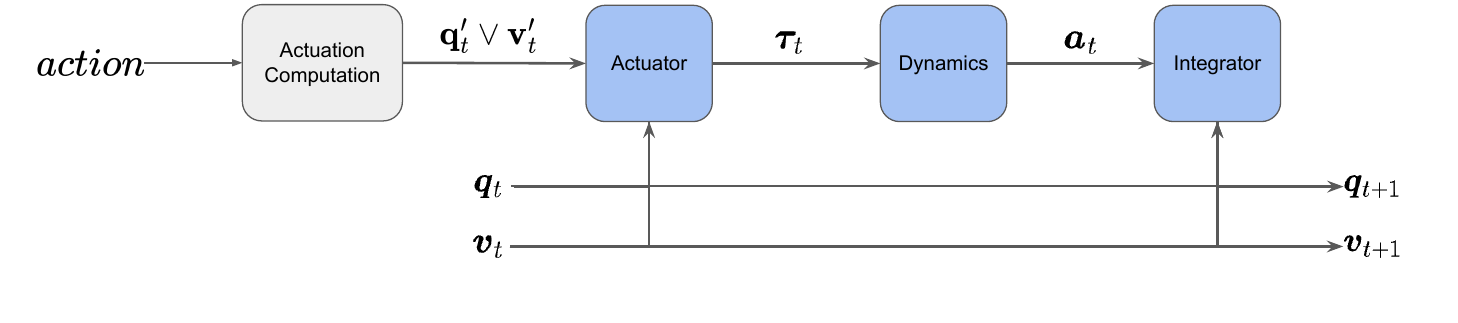}
  \caption{MuJoCo state transition. $action$ is the action produced by the policy. $\boldsymbol{q'}_t$ and $\boldsymbol{v'}_t$ are the target joint positions and velocities, respectively.  $\boldsymbol{q}_t$ and $\boldsymbol{v}_t$ are the current joint positions and velocities. $\boldsymbol{\tau}_t$ is the applied torque and $\boldsymbol{a}_t$ is the current acceleration. $\boldsymbol{q}_{t+1}$, $\boldsymbol{v}_{t+1}$ are the updated joint positions and velocities.}
  \label{fig:mujoco}
\end{figure}

We utilize MuJoCo \cite{todorov2012mujoco} to train our policies in simulation. MuJoCo is a torque-based simulator that faithfully captures real-world dynamics through three important computational functions that are used sequentially: actuator, dynamics, and integrator. Actuators are functions that map actuation to torque ($\bs{\tau}$), the dynamics function maps from applied torque to acceleration, and the integrator updates the state of the simulator by computing next positions and velocities (Fig.~\ref{fig:mujoco}).

\textbf{Actuator.} In our work, we consider two actuation models that are described as follows: 
\begin{align}
f^{ac}_{pos} &:= \bs{K}_p (\q' - \q) - \bs{K}_v \ve,  \text{ (position actuator)} \label{eq:pos_act_sim} \\[3pt]
f^{ac}_{vel} &:= \bs{K}_v (\ve' - \ve),  \text{ (velocity actuator)},
\end{align}
where $\bs{K}_p$ denotes the diagonal matrix of proportional gains or stiffness, and $\bs{K}_v$ represents the diagonal matrix of damping coefficients. Here, $\q'$ and $\ve'$ are the target joint position and velocity, while $\q$ and $\ve$ are the current joint position and velocity. $f_{pos}^{ac}$ is a PD-controller with target velocity $\ve'=0$ and $f_{vel}^{ac}$ is a P-controller.

\textbf{Dynamics Function.}
The dynamics function is then used to compute joint accelerations $\boldsymbol{a}_t$ from applied torque and other forces in the system using the following equation:
\begin{equation}
\label{eq:dynamics}
    \boldsymbol{a}_t = \boldsymbol{M}^{-1}(\boldsymbol{\tau} + \boldsymbol{J}^\top \boldsymbol{f} - \boldsymbol{c}),
\end{equation}
where $\boldsymbol{M}$ is inertia in joint space, $\boldsymbol{\tau}$ is the applied force consisting of actuator torque, passive force, and external forces, $\boldsymbol{f}$ is the constraint force, and $\boldsymbol{c}$ is the bias force consisting of Coriolis, centrifugal, and gravitational forces. $\boldsymbol{J}^T$ maps forces from the constraint space of the simulator to the joint space of the robot.

\textbf{Integrator.}
MuJoCo advances the simulator from time step $t$ to $t{+}1$ using one of several numerical integrators (semi-implicit Euler, implicit-in-velocity, fast implicit-in-velocity, or RK4). In all experiments, we use the fast implicit-in-velocity integrator, which provides a practical trade-off between stability and computational cost \cite{todorov2012mujoco}.

This integrator performs an implicit velocity update that can be written as
\begin{equation}
\ve_{t+1} \;=\; \ve_t \;+\; \delta t \,\widehat{\mathbf{M}}^{-1}\mathbf{M}\mathbf{a}_t,
\qquad
\widehat{\mathbf{M}} := \mathbf{M} - \delta t\,\mathbf{D},
\label{eq:fast-implicit-update}
\end{equation}
where $\mathbf{M}$ is the joint-space inertia matrix and $\mathbf{D} := \partial \mathbf{a}/\partial \ve$. Under the fast implicit-in-velocity integration scheme, the velocity-dependent term is approximated as
$
\mathbf{D} \approx \frac{\partial \boldsymbol{\tau}}{\partial \ve},
$
where $\mathbf{D}$ denotes the Jacobian of the applied torques with respect to joint velocities. In all experiments, we employ MuJoCo’s fast implicit-in-velocity integrator.
 Joint positions are then updated via
\begin{equation}
\q_{t+1} \;=\; \q_t \;+\; \delta t \,\ve_{t+1}.
\end{equation}

\subsection{Real Robot Control System}

For real-world experiments, we control the Franka Emika Panda using the Franka ROS Interface via the Python API wrapper introduced by \cite{saif_sidhik_2020_4320612}. We utilize two actuation interfaces: \texttt{set\_joint\_positions\_velocities} and \texttt{set\_joint\_velocities}.

The \texttt{set\_joint\_positions\_velocities} interface employs a joint impedance controller
\begin{equation}
    f_{\text{imp}}^{ac} := \bs{\tau}_c + \bs{K}_p (\q' - \q) + \bs{K}_v (\ve' - \ve),
\end{equation}
where \(\bs{\tau}_c\) denotes the Coriolis torque, $\q$ and \(\q'\) are the current and target joint positions, \(\ve\) and \(\ve'\) are the current and target joint velocities, and \(\bs{K}_p\) and \(\bs{K}_v\) are diagonal matrices of stiffness and damping gains. Setting \(\ve' = 0\) yields
\begin{equation}
    f_{\text{pos}}^{ac} := \bs{\tau}_c + \bs{K}_p (\q' - \q) - \bs{K}_v \ve,
    \label{eq:pos-act-real}
\end{equation} 
which closely matches the PD position controller used in simulation (Eq.~\ref{eq:pos_act_sim}). The main difference is the omission of $\bs{\tau}_c$, which is implicitly handled by the dynamics function in MuJoCo (Eq.~\ref{eq:dynamics}).

The \texttt{set\_joint\_velocities} interface commands target joint velocities directly. Although its actuator model is not explicitly documented, it can be inferred to correspond to velocity  P-control of the form
\(
    f^{ac}_{\text{vel}} := \bs{\tau}_c + \bs{K}_v (\ve' - \ve).
\) 

The computed torque values are then mapped to electrical currents used to move the joint servos of the physical robot. 

\begin{figure}[t]
    \centering
    
    % --- Row 1 ---
    \includegraphics[width=0.48\columnwidth]{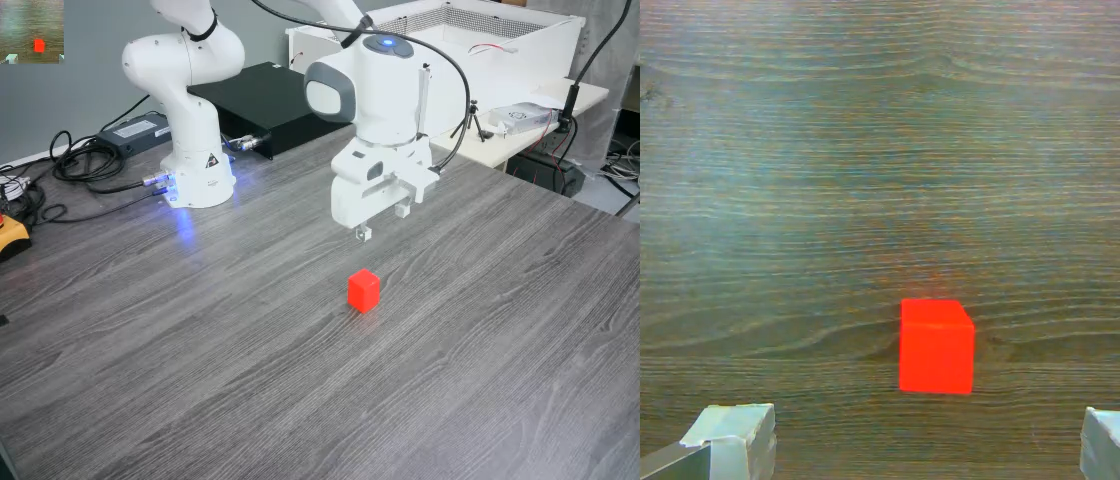}
    \hfill
    \includegraphics[width=0.48\columnwidth]{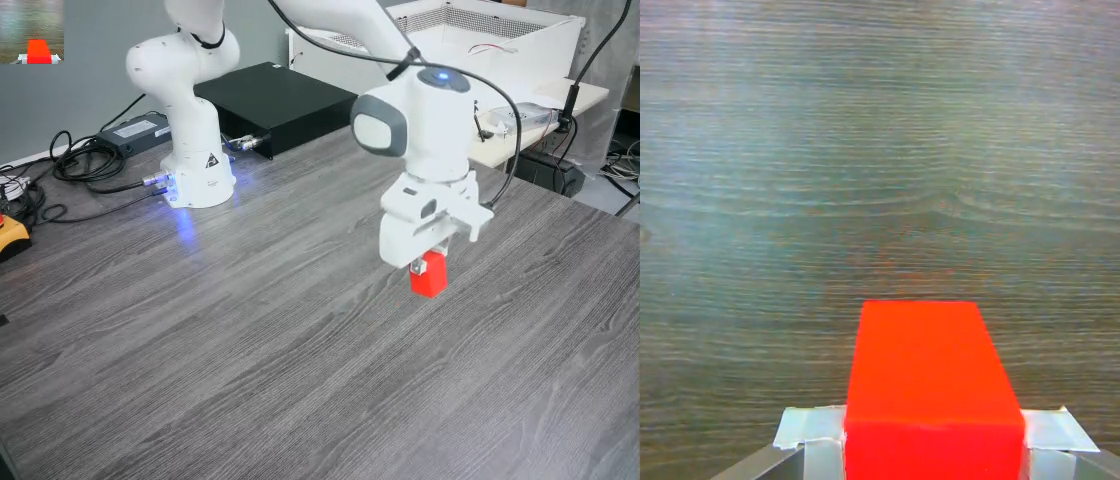}
    
    \vspace{1mm}
    
    % --- Row 2 ---
    \includegraphics[width=0.48\columnwidth]{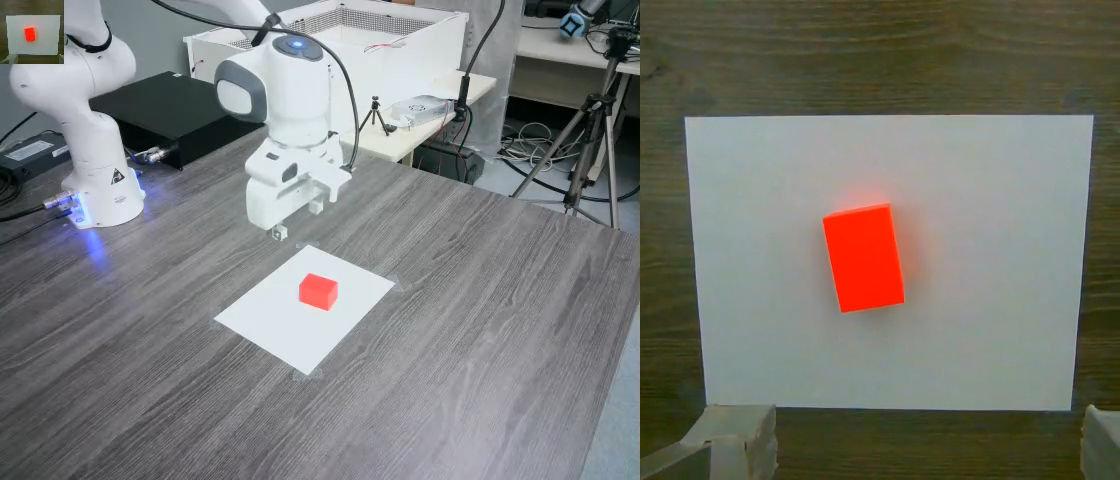}
    \hfill
    \includegraphics[width=0.48\columnwidth]{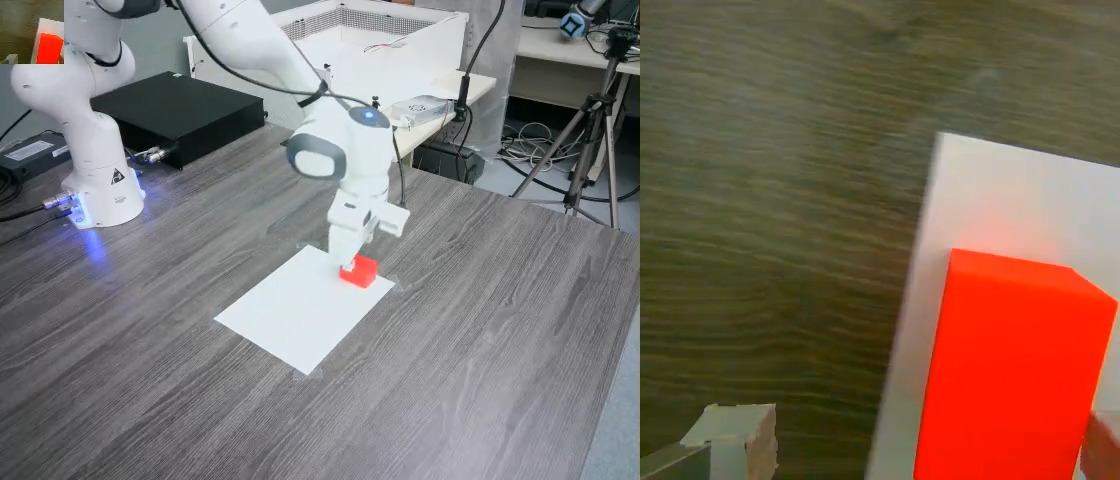}
    
    \caption{Top row: PandaPickCuboid. Bottom row: PandaPushCuboid.}
    \label{fig:real_frames}
\end{figure}

\section{Tasks}
We evaluate our action spaces on two tasks: \textit{PandaPickCuboid} and \textit{PandaPushCuboid}. 
PandaPickCuboid is a prehensile manipulation task in which the robot must reach, grasp, and lift a red cuboid to a minimum height of 17\,cm. Object picking is a fundamental manipulation task that requires precise, multi-stage execution of reaching, grasping, and lifting. In simulation, the episode terminates upon successful lift or timeout. In the real world, episodes additionally terminate if the end-effector exits predefined workspace boundaries or exceeds 10 unsuccessful grasp attempts.

PandaPushCuboid is a non-prehensile manipulation task where the agent must push the same red cuboid across a white base (Fig.~\ref{fig:real_frames}) as much as possible, emphasizing directional control and interaction with object friction and mass. Episodes terminate after a 12\,s timeout in both simulation and real-world settings, and additionally if the end-effector exits workspace boundaries in the real world.

\subsection{Reward Functions}
\textbf{PandaPickCuboid Reward.}
The reward function we use to train the robot is as follows:
\begin{align} 
    \label{eq:reward}
    r_t &:= \max \Big\{ 0 \;,\; g_t^{\text{sum}} - \max_{1 \le i \le t-1} g_i^{\text{sum}} \Big\},
\end{align}
where $g_t^{sum} := \sum_{i=1}^4 w_i \cdot g^{(i)}_t$ is a weighted sum of the reward components with $w_i$ being the weight for the $i$th component (Table~\ref{tab:pick_reward}). 
This formulation ensures that if the reward at time step $t$ is lower than the maximum reward observed in previous steps, the reward for that step is set to zero, effectively considering only improvements in the reward. This formulation was introduced by \cite{Petrenko2023DexPBT}, and we adapted our implementation from \cite{zakka2025mujocoplayground}. 

\begin{table}[t]
\centering
\footnotesize
\setlength{\tabcolsep}{3pt}
\renewcommand{\arraystretch}{1.05}
\caption{Reward components for PandaPickCuboid. 
$\delta_{\text{target}}$ denotes the distance between the block and a fixed target position located vertically above its initial pose at 0.2m. 
$\delta_{\text{rot}}$ is the block upright orientation error. 
$\delta_{\text{gripper}}$ is the gripper--block distance. 
$h_{\text{block}}$ is the block height.}
\label{tab:pick_reward}
\begin{tabular}{p{4.6cm} c p{2.2cm}}
\hline
\textbf{Reward} & \textbf{W} & \textbf{Description} \\
\hline

$g^{(1)} = 1 - \tanh\!\big(5(0.9\,\delta_{\text{target}} 
+ 0.1\,\delta_{\text{rot}})\big)$
& 4
& Pose tracking \\

$g^{(2)} = 1 - \tanh\!\big(5\delta_{\text{gripper}}\big)$
& 8
& Gripper-block distance penalty \\

$g^{(3)} = \mathbb{I}[\text{no floor collision}]$
& 0.25
& Collision penalty \\

$g^{(4)} = \mathbb{I}[|h_{\text{block}} - 0.2| < 0.03]$
& 2.0
& Lift success \\

\hline
\end{tabular}
\end{table}

\textbf{PandaPushCuboid Reward.}
We utilize the following reward function to train policies for the \textit{PandaPushCuboid} task:
\begin{equation}
r_t
=
5\,\mathbb{I}^{\text{in}}_t\, r^{\text{disp}}_t
\;-\;
0.5\, r^{\text{dist}}_t
\;-\;
0.1\,\mathbb{I}^{\text{floor}}_t ,
\label{eq:push_reward}
\end{equation}
where $r^{\text{disp}}_t$ denotes the planar displacement of the block at time step $t$. The indicator function $\mathbb{I}^{\text{in}}_t$ returns $1$ if the block remains within the boundaries of the white base and $0$ otherwise; this is to ensure displacement reward is only given when the block is within the boundaries of the base. The term $r^{\text{dist}}_t$ is the Euclidean distance between the block and the end-effector, designed to encourage the agent to reach the block, while $\mathbb{I}^{\text{floor}}_t$ returns $1$ if the end-effector makes contact with the floor and $0$ otherwise.

\subsection{Observation Spaces}
The observation space comprises four components: (1) an RGB image of size $64 \times 64 \times 3$, (2) a proprioception vector containing joint positions $\q$, joint velocities $\ve$, and grasp status, (3) the end-effector height, and (4) the previous action. Joint positions and velocities are linearly scaled to $[-1,1]$ using $\q = 2\frac{\q - \q_{\min}}{\q_{\max} - \q_{\min}} - 1$ and $\ve = 2\frac{\ve - \ve_{\min}}{\ve_{\max} - \ve_{\min}} - 1$. We empirically observed that scaling improved sim-to-real performance. For PandaPushCuboid, the grasp-status signal is omitted, as grasping is not required.

\subsection{Action Spaces}
The Franka Emika Panda has 7 joints, allowing for 7 degrees of freedom. In the PandaPickCuboid task, our action space has dimensionality 8, with the final dimension reserved for controlling the grasping mechanism. In the PandaPushCuboid task, the dimensionality is 7. All action values are bounded to the range (-1,1), and all action spaces have a cycle time of $\delta t = 40 ms$. We consider four action spaces: pose increment ($\delta \x$), pose velocity ($\dot{\x}$), joint velocities ($\ve$), and joint position increments ($\bs{\delta q}$). These action spaces were selected because they are agnostic to the absolute joint and end-effector positions, making them suitable for our partially observable setup and commonly used in robotic manipulation \cite{zakka2025mujocoplayground, MahmoodKorenkevychKomerBergstra2018, aljalbout2024role}.

\begin{table}[t]
\centering
\small
\setlength{\tabcolsep}{4pt}
\renewcommand{\arraystretch}{1.1}
\caption{Action scaling factors for PandaPickCuboid (Pick)\\ and PandaPushCuboid (Push)}
\label{tab:action_scaling}
\begin{tabular}{l cc cc}
\hline
 & \multicolumn{2}{c}{\textbf{Pick}} & \multicolumn{2}{c}{\textbf{Push}} \\
\cline{2-3} \cline{4-5}
\textbf{Action} & Sim & Real & Sim & Real \\
\hline
Pose inc. ($\delta \x$)      & 0.05 & 0.02 & 0.05 & 0.02 \\
Pose vel. ($\dot{\x}$)       & 0.05 & 0.01 & 0.05 & 0.01 \\
Joint pos. inc. ($\delta \q$)     & 0.05 & 0.06 & 0.05 & 0.05 \\
Joint vel. ($\ve$)                & 1.0  & 0.20 & 1.0  & 0.15 \\
\hline
\end{tabular}
\end{table}

\textbf{Joint Space.}
 Actions are vector outputs from the policy, and the actuator gives them meaning. We define actuators as functions that map actuation to torque. In $\ve$, we utilize velocity actuators and map the target velocities scaled by a constant (c) to torque ($f^{ac}_{vel}: c.\ve^{target} \mapsto \bs{\tau}$) (Table~\ref{tab:action_scaling}). For $\delta \q$, target joint positions are computed as $\q^{target} = \q + c.\delta \q$ and passed to a position actuator ($f_{pos}^{ac}: \q^{target} \mapsto \bs{\tau}$).

\textbf{Cartesian Space.}
For our Cartesian action spaces, the $\delta \x$ representation is actuated using position actuators, whereas $\dot{\x}$ applies incremental commands through velocity actuators. In the $\delta \x$ action space, the first three components correspond to position increments, which specify the desired change in the end-effector’s current position. The next three components correspond to rotational increments about the three Cartesian axes. The seventh component specifies the rotation increment of the robot’s seventh degree of freedom, which is common between all action spaces. Target pose is then computed as $\x^{target} = \x \oplus c.\delta\x$. This target pose is then passed to an analytical inverse kinematics solver proposed by \cite{He2021} and we use the implementation by \cite{zakka2025mujocoplayground}, which maps $\x^{target}$ to target joint positions ($\text{IK}: \x^{target} \mapsto \q^{target}$), then a position actuator is used ($f^{ac}_{pos}:\q^{target} \mapsto \bs{\tau}$). 

The primary difference between $\delta \x$ and $\dot{\x}$ lies in how the resulting joint commands are actuated. For $\dot{\x}$, we reuse the same target joint positions $\q^{target}$ produced by the inverse kinematics solver, but approximate joint velocities using the current joint positions and the action cycle time, given by $\ve^{target} \approx \frac{\q^{target} - \q}{\delta t}$ which then uses a velocity actuator that maps from target velocity to torque ($f^{ac}_{vel}:\ve^{target} \mapsto \bs{\tau}$).

\textbf{Gripper Control}
In PandaPickCuboid, once a policy outputs an action vector $\bs{a}$, the final value of this vector is used to command the gripper into a closed or open state to grasp the object. Gripper activation at a given time step $t$ can be expressed using the following step-wise function: 

\begin{align}
\text{grasp}_t :=
\begin{cases}
1, & \text{if } a_g < \lambda^{1}_g, \\[4pt]
0, & \text{if } a_g \geq \lambda^{2}_g, \\[4pt]
\text{grasp}_{t-1}, & \text{otherwise}.
\end{cases}
\end{align}
where $a_g$ is the scalar gripper action output and $\lambda_g$ is the grasp activation threshold. A value of 1 corresponds to closing the gripper, and a value of 0 corresponds to opening it. During simulation training, we set $\lambda^{1}_g = \lambda^{2}_g = 0$ and we adjusted the thresholds during real-time trials to improve grasping performance by setting them to $\lambda^{1}_g = -0.2$ and $\lambda^{2}_g = 0.9$. In the case of $\delta \x$, we set $\lambda^{1}_g = -0.1$ to improve grasping performance and reduce false grasp attempts.

% \begin{figure*}[t]
%     \centering
%     \includegraphics[width=0.32\textwidth]{figures/results/PandaPickCuboid/training/PandaPickCuboid_eval_episode_reward.pdf}
%     \hfill
%     \includegraphics[width=0.32\textwidth]{figures/results/PandaPickCuboid/training/PandaPickCuboid_eval_episode_success.pdf}
%     \hfill
%     \includegraphics[width=0.32\textwidth]{figures/results/PandaPickCuboid/training/PandaPickCuboid_eval_avg_episode_length.pdf}
%     \caption{PandaPickCuboid simulation training results consisting of episodic return, success rate, and episodic length. 10 independent runs are shown with the median curve highlighted. }
%     \label{fig:pick_sim}
% \end{figure*}

\begin{figure*}[t]
    \centering
    \includegraphics[width=\textwidth, trim=0 1cm 0 0]{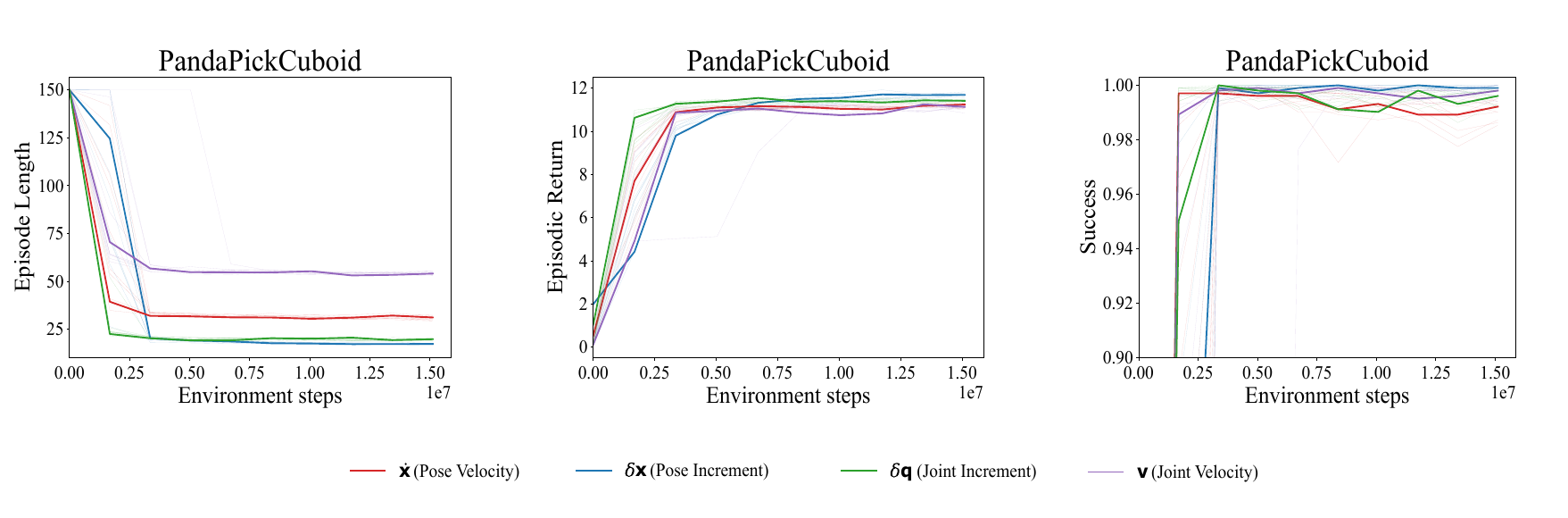}

    \caption{PandaPickCuboid simulation training results consisting of episodic return, success rate, and episodic length. 10 independent runs are shown with the median curve highlighted. }
    \label{fig:pick_sim}
\end{figure*}

\section{Experimental Setup}

In this section, we discuss the real-world setup of the camera and arm, the neural networks, and domain randomization used in our experiments. 
\subsection{Real World Setup}
\begin{figure}[t]
    \centering
    
    \includegraphics[width=0.48\columnwidth]{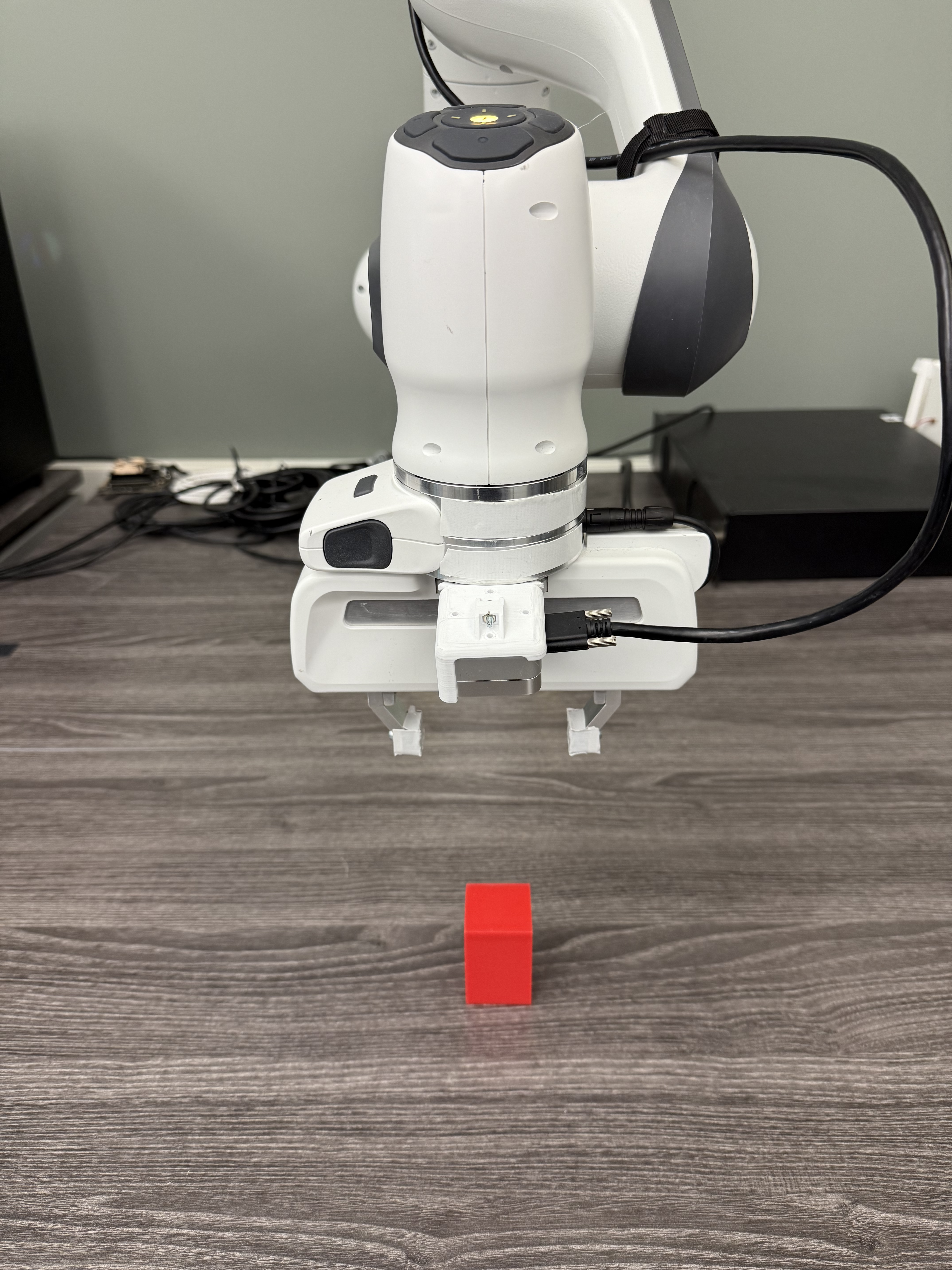}
    \hfill
    \includegraphics[width=0.48\columnwidth]{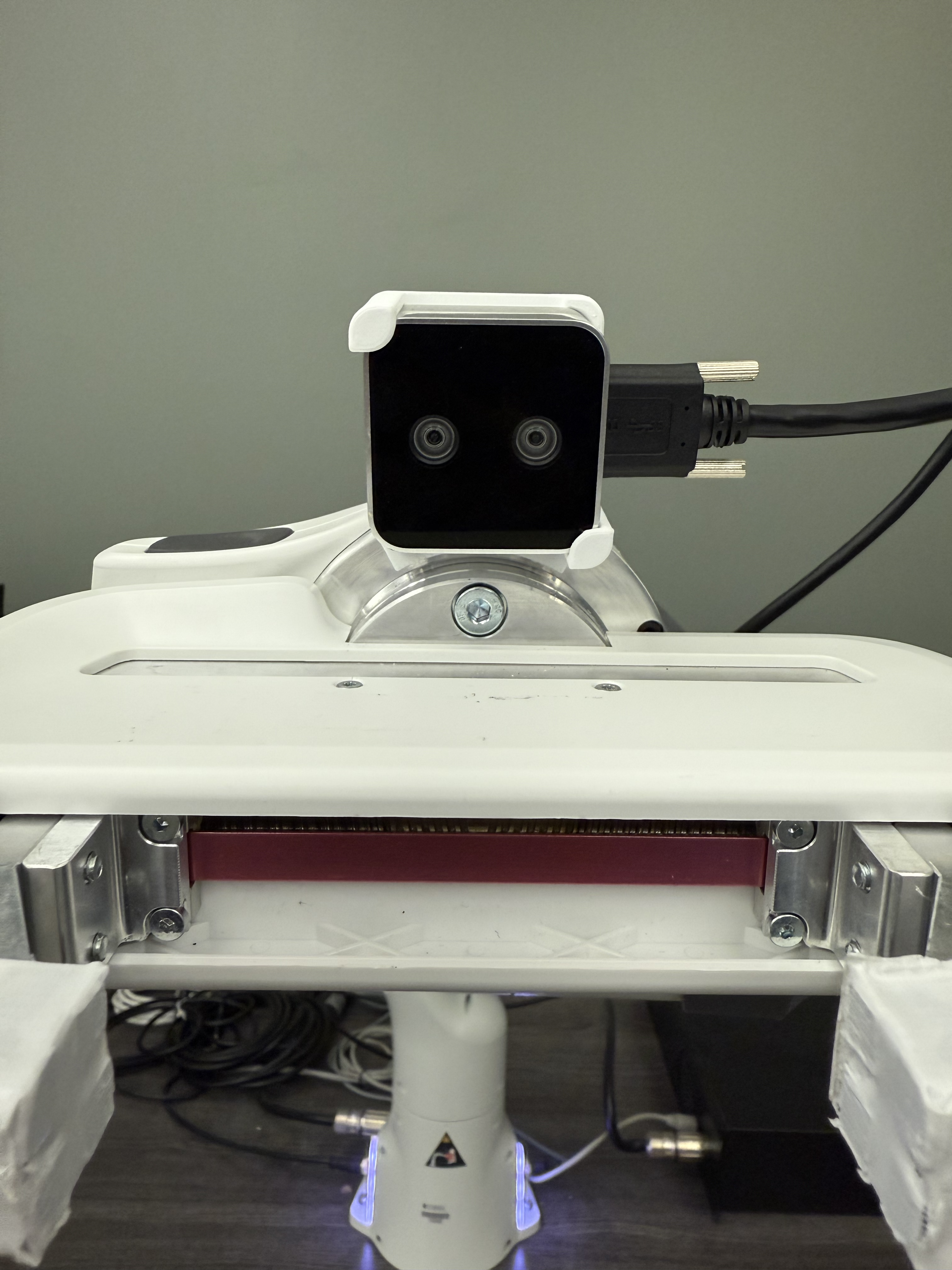}
    
    \caption{Real-world setup and wrist-mounted camera.}
    \label{fig:real_world_setup}
\end{figure}

Our real-world setup consists of a red block with dimensions $4\times4\times6$~cm, an Intel RealSense D405 camera, and the Franka Emika Panda robot arm mounted to the robot's wrist (Fig.~\ref{fig:real_world_setup}). In all real-world trials, the block is placed well within the arm's dexterous workspace and within the camera's field of view.

\begin{table*}[t]
\centering
\small
\caption{Real-world evaluation results for \textit{PandaPickCuboid} and \textit{PandaPushCuboid}. For \textit{Pick}, success reports mean $\pm$ standard error, and other metrics report median $\pm$ IQR. For \textit{Push}, values are reported as median $\pm$ IQR. $\uparrow$ indicates a higher value is preferred. $\downarrow$ indicates a lower value is preferred. Best performers are bolded.}
\resizebox{0.97\textwidth}{!}{
\begin{tabular}{lcccc|ccc}
\hline
& \multicolumn{4}{c|}{\textbf{PandaPickCuboid}} 
& \multicolumn{3}{c}{\textbf{PandaPushCuboid}} \\
\cline{2-8}
\textbf{Action Space} 
& Success (\%) $\uparrow$ 
& Time (s) $\downarrow$
& Collision $\downarrow$
& Jerk ($m/s^3$) $\downarrow$
& Distance (m) $\uparrow$
& Collision $\downarrow$
& Jerk ($m/s^3$) $\downarrow$\\
\hline

Pose Inc. ($\delta \x$)  
& $41.67 \pm 14.23$ 
& $14.78 \pm 11.38$  
& $0.0 \pm 0.0$ 
& $26.88 \pm 17.96$ 
& $0.050 \pm 0.051$ 
& $\bs{0.0 \pm 0.0}$  
& $60.27 \pm 13.61$ \\

Pose Vel. ($\dot{\x}$) 
& $0.00 \pm 0.00$ 
& $8.98 \pm 17.65$ 
& $0.0 \pm 0.0$ 
& $24.55 \pm 10.33$ 
& $0.000 \pm 0.003$ 
& $2.5 \pm 22.0$ 
& $24.85 \pm 5.47$ \\

Joint Pos. Inc. ($\delta \q$)  
& $\bs{100 \pm 0.00}$ 
& $\bs{2.12 \pm 0.59}$   
& $0.0 \pm 0.0$ 
& $52.59 \pm 15.41$ 
& $0.104 \pm 0.071$ 
& $\bs{0.0 \pm 0.0}$  
& $24.37 \pm 2.38$ \\

Joint Vel. ($\ve$)  
& $\bs{100 \pm 0.00}$ 
& $3.58 \pm 0.23$   
& $0.0 \pm 0.0$ 
& $\bs{16.03 \pm 1.34}$ 
& $\bs{0.405 \pm 0.289}$ 
& $8.0 \pm 14.5$ 
& $\bs{14.25 \pm 0.99}$ \\
\hline
Pick Script   
& $\bs{100 \pm 0.00}$ 
& $15.91 \pm 1.57$   
& $0.0 \pm 0.0$ 
& $\bs{11.43 \pm 0.88}$ 
& -- 
& -- 
& --  \\

\hline
\end{tabular}
}

\label{tab:real_world_combined}
\end{table*}

\subsection{Neural Networks}

We employ vision-based actor–critic networks that jointly process RGB image observations and vector-valued observations. Visual features from the image are extracted using a convolutional encoder implemented by \cite{brax2021github} composed of three convolutional layers with ReLU activations. The visual feature vector is concatenated with additional non-visual vector-valued observations, including proprioception, the previous action, the end-effector height, and the grasp status (depending on the task: pick vs push). 

The critic network consists of the convolutional encoder and concatenated observations, followed by a fully connected multilayer perceptron (MLP) head that outputs a scalar state-value estimate. The actor network shares the same structure but differs in its output head, where two fully connected layers parameterize a Gaussian policy by predicting the mean ($\mu$) and standard deviation ($\sigma$) of the action distribution.

\subsection{Domain Randomization}

To improve robustness and facilitate sim-to-real transfer, we apply domain randomization~\cite{TobinFongRaySchneiderZarembaAbbeel2017, PengAndrychowiczZarembaAbbeel2017, Chen2022UnderstandingDR}. During training, we randomize both visual and dynamic parameters. Visual perturbations include camera pose, lighting position and intensity, and image-level properties such as contrast and saturation. We also randomize the block’s initial pose and overlay randomized wooden textures to better match the real-world surface. We inject Gaussian noise into joint position and velocity readings and randomize stiffness and damping.

\subsection{Actuator Tuning}
\label{sec:position_tuning}

When using impedance control on the real robot, appropriately tuned stiffness and damping gains are critical for safe, stable operation. Poorly tuned gains can lead to excessive vibrations, unstable motion, and safety-triggered shutdowns. In our experiments, we reduced stiffness to reduce vibrations and prevent erratic motion.

\section{Real-World Results}

\textbf{PandaPickCuboid.} We train each policy in simulation using PPO over 10 independent seeds with 1024 parallel environments. Fig.~\ref{fig:pick_sim} presents 10 deterministic evaluation runs performed during simulation training. Individual runs are displayed for episodic return, success rate, and episode length. Each action space is denoted by a unique color, and the median run is highlighted. All individual runs are displayed to better portray variation in performance \cite{tanaka2026performancevariationdeepreinforcement}.

In simulation, all action spaces achieve comparable final performance, with success rates in the range $[0.98, 1.00]$. The primary distinction lies in the episodic length, which, in this case, is the time required to complete the task. The pose increment action space ($\delta \x$) achieves the shortest episodes, followed by joint position increment ($\delta \q$).

For real-world evaluation, we conduct 12 deterministic trials per action space (Table~\ref{tab:real_world_combined}). For each action space, we select the top two policies with the highest final episodic return and report results from the policy with the best real-world success. A trial is considered successful if the cuboid reaches at least a target height of 17 cm. Trials terminate upon 10 failed grasp attempts, violation of workspace constraints, or a timeout of 30\,s.

Joint velocity ($\ve$) achieves a 100\% success rate with a median completion time of 3.58\,s. It consistently avoids collisions and exhibits the lowest measured jerk among the evaluated action spaces. Joint position increments ($\delta \q$) also achieve a 100\% success rate and the fastest median completion time of 2.12\,s; however, stable real-world deployment required careful impedance tuning. We also tested our scripted pick policy, which achieves a $100\%$ success rate and a median trial time of $15.91$ seconds. 

In contrast, pose increments ($\delta \x$) demonstrate degraded real-world performance. The policy frequently dragged the cuboid toward the robot base before lifting. To mitigate this behavior, we imposed a lower bound of 0.45\,m on the end-effector $\x$ position during evaluation. Despite this modification, the action space achieved a 41.67\% success rate. Pose velocity ($\dot{\x}$) performed worst in real-world trials, often drifting into unstable configurations or exiting the workspace. It did not successfully complete the task in any trial.

Overall, the simulation results did not fully translate to the real-world experiments. In simulation, $\delta \x$ lifted the block more rapidly, whereas on the real robot, it frequently failed to complete the lifting motion. Likewise, $\dot{\x}$ failed in all real-world trials. These findings suggest that our Cartesian action spaces are more sensitive to sim-to-real discrepancies.

A key difference is that the Cartesian actions rely on an inverse kinematics (IK) mapping to generate joint commands. Even when analytical IK is used, this introduces a kinematic transformation layer. Any discrepancies in link parameters between simulation and reality will propagate through the IK mapping, generating inaccurate joint targets. Furthermore, IK-generated joint trajectories can be highly demanding or abrupt—particularly near singularities. While the simulator's idealized actuators can execute these commands, real-world actuators bounded by physical joint friction, latency, and imperfect impedance gains may struggle to track them accurately, leading to amplified errors.

\begin{figure}[t]
    \centering
    \includegraphics[width=0.8\columnwidth, trim= 0 2cm 0 0]{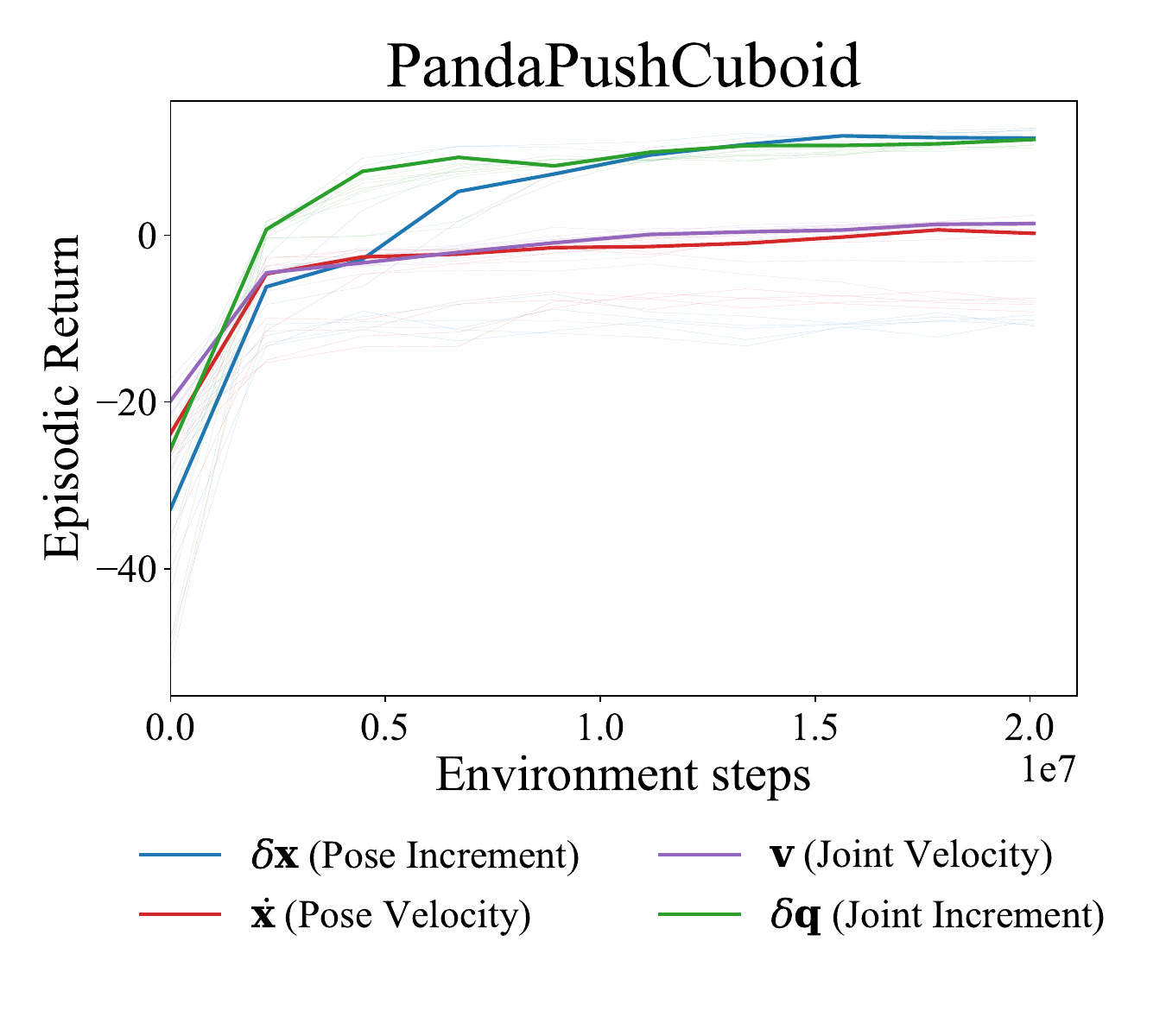}
    \caption{PandaPushCuboid episodic return in simulation training. 10 independent runs are displayed. The median curve is highlighted.}
    \label{fig:push_return}
\end{figure}

\textbf{PandaPushCuboid.} We train each policy in simulation using PPO over 10 independent seeds with 1024 parallel environments. Fig.~\ref{fig:push_return} reports episodic return for 10 deterministic evaluations collected throughout simulation training. All individual runs are displayed. Each action space is represented using a unique color. The median run is highlighted. In simulation, joint position increment ($\delta \q$) and pose increment ($\delta \x$) achieve higher median episodic returns than joint velocity ($\ve$) and pose velocity ($\dot{\x}$), probably due to generating faster trajectories that displace the block more from point-to-point. $\delta \x$ and $\dot{\x}$ exhibit some failed runs, potentially due to the emergence of singularities in the IK solution that stall the robot’s motion. 

For real-world evaluation, we conduct 10 trials per action space using deterministic policies (Table~\ref{tab:real_world_combined}). For each action space, we select the policy with the highest final episodic return and evaluate zero-shot sim-to-real performance. Each trial runs for 12s or terminates if the end-effector exits predefined workspace boundaries that are larger than the white base's boundaries. Aggregate block distance over the trial duration is used as the performance metric.

Joint velocity ($\ve$) achieves the largest median aggregate block distance of 0.405\,m. Although this action space produces the lowest jerk, we observe multiple collisions, primarily due to forceful contact with the cuboid edges. Joint position increments ($\delta \q$) achieve the second-highest displacement (median 0.104\,m). 

In contrast, pose increment ($\delta \x$) and pose velocity ($\dot{\x}$) perform poorly in terms of block distance. However, $\delta \x$ and $\delta \q$ exhibit fewer forceful contact events and appear more compliant during interaction. This compliance does not arise from explicit contact avoidance, but rather from the tuned joint-impedance controller used for position-based actuation, which reduces stiffness and attenuates excessive forces during contact.

% \begin{figure*}[t]
%     \centering
%     \includegraphics[width=0.24\textwidth]{figures/results/PandaPickCuboid/training/PandaPickCuboid_eval_episode_floor_collision.pdf}
%     \includegraphics[width=0.24\textwidth]{figures/results/PandaPushCuboid/training/PandaPushCuboid_eval_episode_floor_collision.pdf}
%     \includegraphics[width=0.24\textwidth]{figures/results/PandaPickCuboid/training/PandaPickCuboid_eval_episode_jerk_per_step.pdf}
%     \includegraphics[width=0.24\textwidth]{figures/results/PandaPushCuboid/training/PandaPushCuboid_eval_episode_jerk_per_step.pdf}
%     \caption{Simulation training metrics: floor collisions and jerk per step for picking and pushing tasks. 10 independent runs are displayed, and the median curve is highlighted for each action space. }
%     \label{fig:training_collision_jerk}
% \end{figure*}

\begin{figure*}[t]
    \centering
    \includegraphics[width=\textwidth]{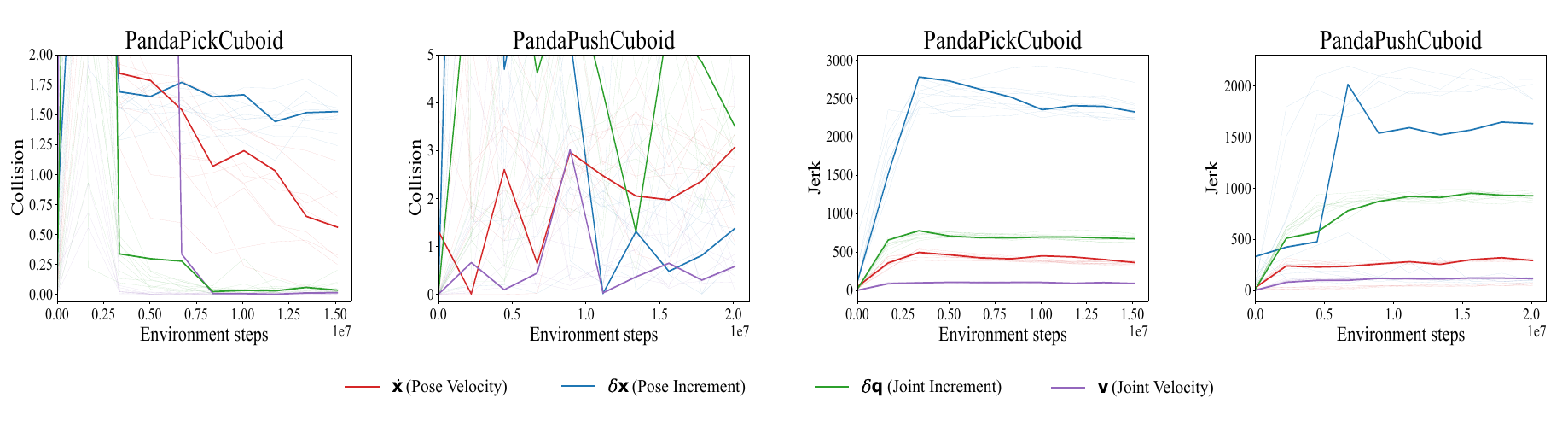}
    \caption{Simulation training metrics: floor collisions and jerk per step for picking and pushing tasks. 10 independent runs are displayed, and the median curve is highlighted for each action space. }
    \label{fig:training_collision_jerk}
\end{figure*}

\section{Training Results on Jerk and Collision}

Figure~\ref{fig:training_collision_jerk} reports episodic floor collisions and average jerk per time step during simulation training, averaged over 1024 parallel environments. These results can inform us of what can happen during learning in the real world. A collision is registered when the gripper body contacts the floor in simulation. Across both PandaPickCuboid and PandaPushCuboid, joint velocity yields the lowest floor-collision counts at convergence, exhibiting more precision during simulation training and exhibiting the smallest jerk.

We approximate jerk at time step $t$ using a finite difference of acceleration of consecutive time steps,
\(
J_t \approx || \frac{\boldsymbol{a}_t - \boldsymbol{a}_{t-1}}{\delta t}||_{2}.
\)
Since joint acceleration is proportional to the applied torque, jerk reflects step-to-step changes in torque. Consequently, larger torque variations across consecutive time steps lead to larger observed jerk. We will now provide a plausible explanation for why joint velocity has the lowest jerk.

 Under position actuation, torque is generated by a position actuator acting on the position error. The resulting torque variation depends on changes in both the position error and joint velocity. In particular, fluctuations in the desired position directly perturb the position error, which is scaled by the proportional gains $\bs{K}_p$. When $\bs{K}_p$ values are large, variations in the commanded position can produce substantial torque changes and, therefore, larger jerk. Under velocity actuation, torque is instead driven by the velocity tracking error and scaled by the velocity gain $\bs{K}_v$. In our experiments, $k_p > k_v$ for all joints. Hence, for comparable command variations, position control can induce larger torque differences than joint velocity. Additionally, with position actuation, the policy directly specifies a desired position (or position increment), so high-frequency fluctuations in the policy output immediately affect the position error and are amplified by the proportional gain. In contrast, velocity actuation produces target velocities, and joint positions evolve via integration,
\(
\q_{t+1} = \q_t + \ve'_t \delta t.
\)
This integration acts as a low-pass filter: high-frequency components in the velocity command are attenuated when mapped to position, introducing temporal smoothing.

\section{Conclusion}
In this work, we benchmarked four commonly used action space representations in vision-based robotic manipulation: joint position increments ($\delta \q$), joint velocity ($\ve$), pose increments ($\delta \x$), and pose velocity ($\dot{\x}$). Policies were trained in simulation on two vision-based tasks, PandaPickCuboid and PandaPushCuboid, and deployed on a physical Franka Emika Panda using sim-to-real transfer.

Across both tasks, joint-space action spaces demonstrated better real-world performance. In particular, joint velocity ($\ve$) achieved the strongest zero-shot sim-to-real results, reaching a $100\%$ success rate in PandaPickCuboid and the highest cuboid displacement in PandaPushCuboid. Joint position increments ($\delta \q$) achieved comparable picking performance, although stable deployment required impedance tuning.

In terms of safety and compliance, action spaces operating through the tuned impedance controller ($\delta \q$ and $\delta \x$) exhibited more compliant behavior and reduced collision severity during contact. However, joint velocity produced the smoothest trajectories overall, achieving the lowest measured jerk and yielding the most natural motion profiles. These results suggest that joint velocity provides a strong balance of robustness, smoothness, and sim-to-real performance. 

\section{Limitation and Future Work}

A limitation of this work is that we consider single-object manipulation tasks involving a single cuboid. Extending the proposed sim-to-real pipeline to diverse objects with varying geometries, sizes, and appearances, as well as to multi-object or cluttered scenes, and complex manipulation tasks like screwing, hammering, and manipulating objects with significant variation in mass and friction, remains future work. In particular, object selection in cluttered environments presents a scalability challenge. One possible direction is to incorporate object detection to define regions of interest (ROIs) for target selection \cite{TobinFongRaySchneiderZarembaAbbeel2017}. Alternatively, mask-based goal conditioning and ROI selection could enable specification and manipulation of target objects in cluttered scenes \cite{shahriar2025generalefficientvisualgoalconditioned}.

This paper considers four commonly used action spaces for robots; there are more, such as joint position or torque control.
We did in fact experiment with torque-based control, but found it was more difficult to train and led to unstable behavior in simulation. 
In real-world deployment, torque control required explicit gravity and Coriolis compensation, complicating stable lifting during the picking task. A systematic evaluation of torque control for contact-rich manipulation is left for future investigation.

Finally, this study focuses on object manipulation; a line of future work is to perform a similar action-space comparison in other classes of robotic tasks, such as locomotion and other embodiments, such as drones and humanoid robots. 

\section*{Acknowledgement}
We would like to thank anonymous reviewers for their constructive comments and feedback, which helped improve this work. We would like to thank Shivam Garg, Mustafa Heidarbhai, and Haruto Tanaka for helpful discussions. This research was supported in part by the National Research Council Canada (NRC) and by the CIFAR AI Chair program. We are also appreciative of the computing resources provided by the Digital
Research Alliance of Canada and the financial support from the RLAI
laboratory and Amii.

\bibliographystyle{IEEEtran}

\bibliography{ref}

\end{document}